\documentclass[10pt, a4paper]{article}
\usepackage{lrec2022} 
\usepackage{multibib}
\newcites{languageresource}{Language Resources}
\usepackage{graphicx}
\usepackage{tabularx}
\usepackage{soul}
\usepackage{titlesec}
\titleformat{\section}{\normalfont\large\bfseries\center}{\thesection.}{1em}{}
\titleformat{\subsection}{\normalfont\SmallTitleFont\bfseries\raggedright}{\thesubsection.}{1em}{}
\titleformat{\subsubsection}{\normalfont\normalsize\bfseries\raggedright}{\thesubsubsection.}{1em}{}
\renewcommand\thesection{\arabic{section}}
\renewcommand\thesubsection{\thesection.\arabic{subsection}}
\renewcommand\thesubsubsection{\thesubsection.\arabic{subsubsection}}

\usepackage{epstopdf}
\usepackage[utf8]{inputenc}

\usepackage[hidelinks]{hyperref}
\usepackage{xstring}

\usepackage{color}

\usepackage{booktabs}
\usepackage{adjustbox}
\usepackage{threeparttable}
\usepackage{lipsum}
\usepackage{float}
\usepackage{multirow}

\title{LIP-RTVE: An Audiovisual Database for Continuous Spanish in the Wild\\}

\name{David Gimeno-Gómez, Carlos-D. Martínez-Hinarejos} 

\address{Pattern Recognition and Human Language Technologies Research Center,\\
         Universitat Politècnica de València, Camino de Vera, s/n, 46022, València, Spain \\
         \{dagigo1, cmartine\}@dsic.upv.es\\}

\abstract{
Speech is considered as a multi-modal process where hearing and vision are two fundamentals pillars. In fact, several studies have demonstrated that the robustness of Automatic Speech Recognition systems can be improved when audio and visual cues are combined to represent the nature of speech. In addition, Visual Speech Recognition, an open research problem whose purpose is to interpret speech by reading the lips of the speaker, has been a focus of interest in the last decades. Nevertheless, in order to estimate these systems in the currently Deep Learning era, large-scale databases are required. On the other hand, while most of these databases are dedicated to English, other languages lack sufficient resources. Thus, this paper presents a semi-automatically annotated audiovisual database to deal with unconstrained natural Spanish, providing 13 hours of data extracted from Spanish television. Furthermore, baseline results for both speaker-dependent and speaker-independent scenarios are reported using Hidden Markov Models, a traditional paradigm that has been widely used in the field of Speech Technologies. 
 \\ \newline  }

\begin{document}
\maketitleabstract

\section{Introduction} \label{intro}

\begin{table*}[!hbtp]
\begin{adjustbox}{width=\textwidth}
\begin{threeparttable}
\begin{tabular}{cccc}
\toprule
\multicolumn{4}{c}{\textbf{LIP-RTVE: an Audiovisual Database for Continuous Spanish in the Wild}} \\ \midrule
\textbf{Audio Resolution} & 16 kHz mono-channel & 16 bit-depth & WAV format \\
\textbf{Video Resolution} & 25 frames/second & RGB images & PNG format \\
\textbf{Duration} & $\sim$13 hours & 10,352 overlapped samples & 1,168,087 frames \\
\textbf{Speakers} & \textbf{Total:} 323 & \textbf{Males:} 163 & \textbf{Females:} 160 \\
\textbf{ROIs Average Size} & \textbf{FitMouth:} 27$\times$16 pixels & \textbf{WideMouth:} 45$\times$30 pixels & \textbf{FaceROI:} 55$\times$58 pixels \\
\textbf{Vocabulary} & 9308 unique words & \multicolumn{2}{c}{\textbf{Running Words:} 140,123 words} \\
\textbf{Phonemes} & 24 unique phonemes & \multicolumn{2}{c}{\textbf{Running Phonemes:} 654,368 phonemes} \\
\textbf{Characters} & 28 unique characters & \multicolumn{2}{c}{\textbf{Running Characters:} 801,830 characters} \\
\textbf{Speech Rate (words/second)} & \textbf{Min:} 0.58 & \textbf{Median:} 2.94 & \textbf{Max:} 9.73 \\
\textbf{Words per Utterance} & \textbf{Min:} 1 & \textbf{Median:} 12 & \textbf{Max:} 62 \\
\textbf{Phonemes per Utterance} & \textbf{Min:} 4 & \textbf{Median:} 55 & \textbf{Max:} 270 \\
\textbf{Characters per Utterance} & \textbf{Min:} 4 & \textbf{Median:} 68 & \textbf{Max:} 343 \\
\textbf{Seconds per Utterance} & \textbf{Min:} 0.97 & \textbf{Median:} 4.00 & \textbf{Max:} 15.97 \\ \bottomrule
\end{tabular}
\end{threeparttable}
\end{adjustbox}
\caption{Overall details regarding the compiled LIP-RTVE Audiovisual Database.}
\label{overall}
\end{table*}

Despite the fact that speech perception is commonly considered as a purely auditory process, the truth is that it is a process involving multiple senses, as well as high-level knowledge related with grammar and semantics \cite{dupont2000}. In fact, the study carried out by \newcite{01mcgurk1976hearing} demonstrated the importance of visual information and its relationship with the sounds produced. Nevertheless, for deaf or hearing-impaired people, who are totally or partially dependent on their sense of sight, speech understanding poses a great challenge since, as \newcite{duchnowski2000development} support, only 30\% of speech information is visible. For this reason, different areas of research have focused their efforts on speech recognition when the auditory sense is not functional, such as lipreading \cite{kaplan1987speechreading,rodriguez2008lipreading}, cued speech \cite{cornett1967cued}, or silent speech interfaces \cite{denby2010silent}.

Regarding the field of Speech Technologies in its origins, Automatic Speech Recognition (ASR) systems were focused only on processing the acoustic signal. Nowadays, this type of systems reaches high-quality performances \cite{chan2015listen}. However, these approaches suffer a deterioration in quality when the audio signal is damaged or corrupted \cite{juang1991adverse}. As a consequence, in order to deal with this issue, the research was impulsed towards Audio-Visual Speech Recognition (AVSR) approaches \cite{potamianos2003recent,dupont2000}. In this way, as the above mentioned studies demonstrate, it was shown that the combination of acoustic and visual cues could represent the nature of speech more robustly. On the other hand, as \newcite{fernandez2018survey} present, in the last decades there has been an increasing interest in the Visual Speech Recognition (VSR) task, an open research problem where the automatic system aims to interpret speech by reading the lips of the speaker. In fact, the current state-of-the-art in this challenging task has been achieved thanks to the incorporation of acoustic cues during the training phase \cite{ma2022visual,afouras2018deep}.

Notwithstanding, it is well-known that data is a fundamental pillar for this area of research. Thus, two relevant aspects must be taken in consideration. First, in the currently Deep Learning (DL) era, large-scale audiovisual databases are required in order to estimate the immense amount of parameters that form this type of systems. On the other hand, there is a language unbalanced proportion, since while most of these databases are dedicated to English, other languages lack of sufficient resources \cite{fernandez2018survey,zadeh2020moseas}.

\textbf{Contributions:} due to the above presented reasons, this paper presents an audiovisual database to deal with unconstrained natural Spanish, following the so-called \textsl{in the wild} philosophy. More precisely, around 13 hours of data extracted from Spanish broadcast television have been semi-automatically collected. In this way, we intended to ensure the quality of the compiled data. On the other hand, baseline results for both speaker-dependent and speaker-independent scenarios are reported using Hidden Markov Models (HMMs), a traditional paradigm that has been widely used in the field of Speech Technologies \cite{gales2008application}. Furthermore, different input modalities, i.e., acoustic, visual or audiovisual features, have been studied. In fact, an audio-only approach was considered as the lower bound for our proposed task.

\section{Related Work} \label{related}

As \newcite{fernandez2018survey} suggest, advances achieved in the field of audiovisual Speech Technologies have been conditioned, among other reasons, by the available audiovisual databases at the time. In its origins, these databases began by collecting data in order to deal with simple tasks like alphabet or digit recognition, such as AVLetters \citelanguageresource{matthews2002extraction} and CUAVE \citelanguageresource{patterson2002cuave} corpora, respectively. Nonetheless, in the last decade numerous large-scale publicly available audiovisual databases which address natural speech recognition have been compiled \cite{fernandez2018survey}. Thus, due to the nature of our contribution, this section is focused on this type of databases, with special emphasis on those that follow the so-called \textsl{in the wild} philosophy.

With respect to databases that have been recorded in a controlled setting, the 4-hours RM-3000 \citelanguageresource{howell2016visual} database is focused on natural speech, but the main inconvenient is that this corpus only has one speaker. Another database we have to mention, despite having also been recorded in controlled conditions, is the TCD-TIMIT \citelanguageresource{harte2015tcd} corpus, which offers around 7 hours of data collected from 62 different speakers. On the other hand, in relation with realistic scenarios, we must mention the LRS2-BBC \citelanguageresource{lrchung2017lip,afouras2018deep}, MV-LRS \citelanguageresource{chung2017profile}, and LRS3-TED \citelanguageresource{afourasLRS3data} corpora, all of them automatically collected from mass media. In this way, each one of these databases offers hundreds of recorded hours, providing an adequate support for training architectures based on DL techniques. However, a noteworthy fact is that all these resources are dedicated to English. 

Regarding Spanish, our language of interest, there has been an increase in available resources in the last years. Nevertheless, it is not comparable with the vast amount of data mentioned above. First, we must mention the VLRF \citelanguageresource{fernandez2017towards} corpus, where 25 speakers provide around 3 hours of natural sentences but recorded in controlled conditions. Furthermore, speakers were asked to strive to vocalize in an appropriate and expressive way. On the other hand, the recent multilingual CMU-MOSEAS \citelanguageresource{lrzadeh2020moseas} database covers 4 low-resources languages, including Spanish. Concretely, although a large amount of data was compiled, only about 18 hours of samples were annotated for each one of these languages. Additionally, this is an interesting corpus, as it provides a multi-modal point of view, supplying information related with the emotions and subjectivity expressed by the speaker. Finally, \newcite{diana2019audiovisual} defined, inspired by the large-scale English-based corpora previously mentioned, a process to automatically collect audiovisual data from Youtube videos. In this way, a database with around 100,000 annotated samples and the employed automatic collector software were made publicly available.

\section{LIP-RTVE Database}

The compiled audiovisual database is composed of around 13 hours of semi-automatically collected and annotated data, whose main overall details and statistics are depicted in Table \ref{overall}.

Nonetheless, as it is reflected along this section, it is necessary to mention that the LIP-RTVE database was conceived at the first instance as a corpus focused on the Automatic Lipreading or VSR task. Thus, our purpose was to increase the available language resources to support the research regarding VSR for unconstrained and natural Spanish in the wild. However, in our experiments (see Section \ref{baseline}), different input modalities were studied, among which the audio-only approach was considered as the lower bound for our proposed task.

\subsection{Source Data} \label{source-data}

In order to provide an appropriate support to estimate robust automatic systems against realistic scenarios, we decided to extract our corpus from TV broadcast programmes. Thus, we compiled it from a subset of the RTVE database \citelanguageresource{11lleida2018rtve2018} which has been employed in the Albayzín evaluations \cite{12lleida2019albayzin}. Concretely, despite the fact that this database is made up of different programmes broadcast by \textsl{Radio Televisión Española}, we compiled our corpus only from the news programme 20H. 

Thereby, the corpus belongs to the so-called \textsl{in the wild} philosophy, offering a large number of speakers in a wide range of scenarios, either inside a record studio or in outdoor locations, where the speaker does not always maintain a frontal plane but can sometimes adopt tilted postures. Furthermore, it includes variations on intra-personal aspects, light conditions, or in distance from the speaker to the camera. It is remarkable that not all the compiled speakers are well-trained television professionals, but a considerable number of them are interviewees who speak naturally, making mistakes or hesitating. In fact, these and other types of spontaneous speech phenomena, as it is described in Section \ref{challenges}, were identified throughout the entire database.

On the other hand, regarding the details of the recording setting, this subset contains MP4-format files with a 48 kHz two-channel stereo audio resolution and videos recorded with a resolution of 480$\times$270 pixels at 25 fps.

One aspect that must be noted is that the RTVE database is protected by a Non-Disclouse Agreement (NDA)\footnote{\url{http://catedrartve.unizar.es/rtvedatabase.html}}, which implies the signing of this license to be able to access our source data. In any case, this license allows to freely use the audiovisual material for research purposes.

\subsection{Methodology} \label{method}

\begin{table*}[!hbtp]
\begin{adjustbox}{width=\textwidth}
\begin{threeparttable}
\begin{tabular}{ccccccccccc}
\toprule
 \multicolumn{2}{c}{\multirow{2}{*}{\textbf{Dataset}}} & \multirow{2}{*}{\textbf{Duration}} & \multicolumn{3}{c}{\textbf{Speakers}} & \multirow{2}{*}{\textbf{Utterances}} & \multirow{2}{*}{\parbox{1.5cm}{\centering \textbf{Running \\ Words}}} & \multirow{2}{*}{\textbf{Vocabulary}} & \multicolumn{2}{c}{\textbf{Language Model}} \\ \cmidrule{4-6} \cmidrule{10-11}
 &  &  & \textbf{Males} & \textbf{Female} & \textbf{Total} & & & & \textbf{Perplexity} & \textbf{OOV words}  \\ \midrule
\multicolumn{1}{c}{\multirow{3}{*}{\textbf{SI}}} & \textbf{TRAIN}  & $\sim$9 hours & 10 & 19 & 29 & 7142 & 99449 & 7524 & 98.9 & 755 \\ 
\multicolumn{1}{c}{} & \textbf{DEV}  & $\sim$2 hours & 86 & 65 & 151 & 1638 & 20541 & 2932 & 107.1 & 191 \\ 
\multicolumn{1}{c}{} & \textbf{TEST} & $\sim$2 hours & 67 & 76 & 143 & 1572 & 20133 & 2983 & 104.2 & 193 \\ \midrule
\multicolumn{1}{c}{\multirow{3}{*}{\textbf{SD}}} & \textbf{TRAIN} & $\sim$9 hours & 163 & 160 & 323 & 7355 & 96174 & 8244 & 100.5 & 782 \\ 
\multicolumn{1}{c}{} & \textbf{DEV} & $\sim$2 hours & 100 & 119 & 219 & 1597 & 22670 & 4316 & 98.5 & 192 \\ 
\multicolumn{1}{c}{} & \textbf{TEST} & $\sim$2 hours & 55 & 68 & 123 & 1400 & 21259 & 4133 & 105.4 & 165 \\ \bottomrule
\end{tabular}
\end{threeparttable}
\end{adjustbox}
\caption{Details regarding the training (TRAIN), development (DEV), and testing (TEST) data sets defined in LIP-RTVE for both speaker-independent (SI) and speaker-dependent (SD) scenarios. For each data set, the perplexity and the number of Out Of Vocabulary (OOV) words were computed based on the language model described in Section \ref{lms}.}
\label{datasets}
\end{table*}

The MP4-format files provided by the source data had to be pre-processed since, on numerous occasions, voice-over was used or more than one speaker appeared on the scene, aspects that were not suitable to deal with the VSR task. Therefore, we defined a methodology to obtain samples that were appropriate for both ASR and VSR at the same time.

For this reason, in order to ease the collecting process, our first step was to implement an automatic software to obtain extracts from the MP4 files where at least one face appeared on the scene. This stage was made possible thanks to the use of the face detection tools described in Section \ref{rois}. Then, once these extracts were obtained, we selected those where a unique speaker was talking for a maximum of 15 seconds. Nevertheless, those scenes where other people appeared in the background did not pose a problem and were accepted as new samples of the database, since the Region of Interest (ROI) extraction process was implemented to capture the face that occupies the largest area in the scene. This process was manually supervised, since in certain situations the largest face did not always correspond to the person speaking.

Subsequently, one aspect which must be mentioned is that we split each long sample into smaller ones, as long as the speaker made pauses in his or her speech that allowed us to make an adequate division of the message. In this way, at expense of building a corpus with overlap, we were able to increase the amount of available data. 

Finally, each sample of the database was manually annotated, obtaining its corresponding transcription. The pre-processing details regarding these transcriptions are described in Section \ref{trans}.

\subsection{Region of Interest Extraction} \label{rois}

When facing VSR it is necessary to apply Computer Vision techniques in order to extract our ROIs. In this case, we are talking about the face of the speaker, a region where it is contained the information related with face expressions that would allow us to address the lipreading task. Thus, by using open-source resources\footnote{\url{https://github.com/hhj1897/face_alignment}}\textsuperscript{,}\footnote{\url{https://github.com/hhj1897/face_detection}} \cite{deng2020retinaface,bulat2017far}, an automatic software to identify 68 facial landmarks \cite{sagonas2016} was implemented and released together with the database. 

\begin{figure}[H]
\centering
\includegraphics[width=0.75\columnwidth]{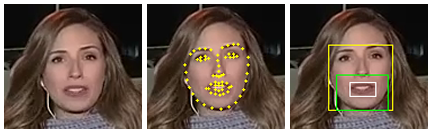}
\caption{The Region of Interest extraction process. White box: \textsl{fitMouth}. Green box: \textsl{wideMouth}. Yellow box: \textsl{faceROI}.}
\label{rois2}
\end{figure}

Once these landmarks were found, by selecting some of them, we were able to define the three types of ROIs depicted in Figure \ref{rois2}. Henceforth, these ROIs, from the smallest to the largest size, are referenced as \textsl{fitMouth}, \textsl{wideMouth}, and \textsl{faceROI}, whose average sizes are reflected in Table \ref{overall}. As we have previously suggested in Section \ref{method}, in each frame the face occupying the largest area on the scene was selected in order to avoid confusion with people in the background.

The reason why we decided to define ROIs with different sizes was due to the differences that exist between the approaches based on end-to-end DL structures \cite{chan2015listen,chung2017lip,afouras2018deep,ma2022visual} and those based on the traditional paradigm of HMMs \cite{gales2008application,thangthai2015improving}. The latter is made up of several modules where each of them is independent from the other. This fact implies that visual speech features will be static during the training phase of the speech module and, therefore, a smaller and specific ROI should be more convenient. Conversely, by employing end-to-end approaches, all their parameters, including those in charge of extracting the visual speech features, are estimated according to the mistakes found during the message decoding phase. In this way, end-to-end approaches are able to identify or select relevant features in a wider ROI which might additionally provide more useful information \cite{zhang2020rois}.

\subsection{Audio Files}

The acoustic signal from each sample was transformed into a 16 kHz mono-channel with 16 bit-precision WAV file, as detailed in Table \ref{overall}. This process was made possible by using the open-source library FFmpeg \cite{tomar2006converting}. In the same way as with the ROIs, a software was released to process the acoustic signals.

\subsection{Transcriptions} \label{trans}

As a pre-processing, after lowercasing all the text, all punctuation marks as well as accents were removed. Finally, transcriptions were coded using the UTF-8 standard.

\subsection{Identified Challenges} \label{challenges}

The database presents all those challenges that could be expected in a realistic scenario, as it is reflected in Table \ref{overall}. Some details we must comment are that, in certain samples, the speaker may speak too quickly, as it is often the case on news programmes. Another feature is that there are samples with considerable differences in length, from samples where we can find numerous words to samples where the speaker only pronounces a word. Additionally, there is a remarkable unbalance between the participation of the each one of speakers in terms of seconds.

Regarding acoustic signals, there are many occasions where we found background noises that could complicate the understanding of the message. Nevertheless, as we have mentioned at the beginning of this section, the LIP-RTVE database was primarily designed to address the VSR task. For this reason, we must highlight the following identified lipreading-related challenges:

\begin{itemize}
    \item Complex silence modelling \cite{thangthai2018computer}. We could consider that the speaker is silent when his or her mouth is closed or when there is no lip movements. The former is not always true and, additionally, there are certain phonemes, such as the sound /p/, that are produced by bringing the lips together. Regarding the lack of lip movements, there are sounds that are mainly produced from the throat with an imperceptible participation of the tongue or lips. 
    
    \item Visual ambiguities, since several phonemes can be associated with one viseme, i.e. the basic speech unit in the video domain \cite{fisher1968confusions}. In other words, there is no one-to-one correspondence between both entities. The clearest example would be the ambiguity that exists when visually discerning between the phonemes /p/, /b/, and /m/.
    
    \item Co-articulation caused by context influence. As
    \newcite{fernandez2017optimizing} suggest in their study, there are phonemes whose visual correspondence can suffer noticeable changes depending on their surrounding context.
    
    \item Finally, certain aspects, such as wetting the lips, poor vocalization, errors and rectifications, or even lowering the head to read notes, could hinder the correct learning of the system in some way.

\end{itemize}

Thus, we must be aware of the challenges that the lack of the auditory sense implies in the automatic speech recognition field.

\subsection{Public Release} \label{publicrelease}

Unfortunately, several details related with the NDA license of the RTVE database must be considered before sharing our contribution with the rest of the research community. For this reason, the entire LIP-RTVE database has not yet been publicly released. Nevertheless, as the data is processed, all the details and resources needed to obtain our database in a license-respecting way will be available on the authors' Github Repository\footnote{\url{https://github.com/david-gimeno/LIP-RTVE}} as soon as possible.

\section{Experimental Setup}

\subsection{Data Sets}
The LIP-RTVE database offers two partitions both for a speaker-dependent (SD) and speaker-independent (SI) scenario. Each partition, in order to define an experimental benchmark, was split in specific training (TRAIN), development (DEV) and testing (TEST) sets. Nonetheless, due to the nature of the source data, there is a significant unbalance regarding the participation of each speaker in the compiled corpus. For this reason, with the intention of providing the best possible learning to the automatic system, the speakers with the longest appearances were allocated first to the training set until reaching the 70\% of the total data. Then, the remaining samples were randomly assigned to the DEV or TEST set, gathering, for each of them, around 15\% of data. The main details regarding these data sets for each scenario are depicted in Table \ref{datasets}. We must mention that in the SD scenario different aspects were taken into account in order to provide a non-overlapping partition.

\subsection{Acoustic Features} \label{mfccs}

The standard representation in the field of ASR was applied on acoustic signals. More specifically, the 39-dimensional Mel Frequency Cepstral Coefficients (MFCC) and their corresponding first- and second-order dynamic differential parameters ($\Delta{+}\Delta\Delta$) \cite{gales2008application} were extracted at 100 fps.

\subsection{Visual Features} \label{visualfeats}

\begin{figure}[H]
\centering
\includegraphics[width=0.8\columnwidth]{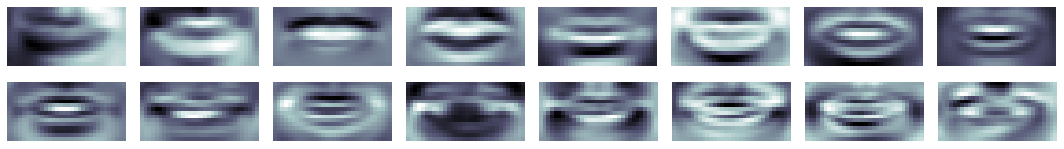}
\caption{The eigelips obtained in the speaker-independent scenario.}
\label{eigen16}
\end{figure}

Unlike acoustic ASR, there is no consensus on which is the best option to represent the nature of visual speech \cite{fernandez2018survey}. Thus, as this concept has been widely studied in the field of VSR \cite{thangthai2018comparing,lan2009comparing}, we decided to extract the appearance-based features known as eigenlips. By selecting 25 random frames from each training sample, we computed the Principal Component Analysis (PCA) technique \cite{wold1987principal}, reducing each frame into 16 components. These eigenlips are shown in Figure \ref{eigen16}, where we can observe how each component focuses on different aspects, such as lip contours or zones where we can find teeth and tongue.

On the other hand, as it was explained in Section \ref{rois}, due to the nature of traditional ASR paradigms, as it is the case of our experiments (see Section \ref{asr}), we considered to use the \textsl{fitMouth} ROIs as the better option to extract the visual speech features. More specifically, all these ROIs were normalized to a resolution of 32$\times$16 pixels, converted to gray-scale images and, in addition, a histogram equalization was applied to them.


\subsection{Automatic Speech Recognition System} \label{asr}

The ASR system employed in our research was designed in the Kaldi toolkit \cite{povey2011kaldi}, where several workflows or recipes to build different paradigms in the field of Speech Technologies are provided. Concretely, we defined a traditional HMM-based system in combination with Gaussian Mixture Models (GMMs) \cite{gales2008application}, taking as a reference the Wall Street Journal (WSJ) recipe\footnote{\url{https://github.com/kaldi-asr/kaldi/tree/master/egs/wsj/s5}}. In order to facilitate the understanding of the results reported in Section \ref{baseline}, we must briefly describe the different stages that compound the estimation process of a GMM-HMM system. In this way, we distinguish the following phases:

\begin{itemize}
    \item \textbf{MONO}: a context-independent GMM-HMM is estimated from scratch applying, over the raw features, the Cepstral Mean and Variance Normalization (CMVN) technique and $\Delta{+}\Delta\Delta$ coefficients \cite{gales2008application}.

    \item \textbf{DELTAS}: in this phase, a context-dependent GMM-HMM is trained, employing a decision tree-based triphone state clustering \cite{young1994tying}. The input features remain identical to the previous step.
    
    \item \textbf{LDA+MLLT}: in this stage, the Linear Discriminant Analysis (LDA) \cite{rao1965linear} and Maximum Likelihood Linear Transform (MLLT) \cite{gopinath1998maximum} techniques are applied to compute the known as HiLDA features \cite{potamianos2001hierarchical}, whose purpose is to reduce the feature dimensionality and capture contextual information. Thus, the GMM-HMM is re-estimated.
    
    \item \textbf{SAT}: the last GMM-HMM is obtained by applying a Speaker Adaptive Training (SAT) \cite{anastasakos1997speaker} based on the feature space Maximum Likelihood Linear Regression (fMLLR) method \cite{gales1998fmllr}.
    
\end{itemize}

Finally, the decoding phase is based on a Weighted Finite-State Transducer \cite{mohri2008speech} which integrates the morphological model, phonetic context-dependencies, the lexicon, and the language model.

\subsection{Lexicon and Language Models} \label{lms}

In order to estimate both models, around 80k sentences were collected from other news programmes broadcast by RTVE during the same period of time. 

In this way, the lexicon model was built, integrating a vocabulary of 45247 unique words. The foundations of this model are based on a vocabulary of 24 phonemes defined according to Spanish phonetic rules \cite{quilis1997principios} in addition to the default \emph{silence} phones of Kaldi.

On the other hand, a 4-gram word-based language model was estimated using the SRLIM toolkit \cite{stolcke2002srilm}. Nonetheless, as it is detailed in Table \ref{datasets}, the high perplexity and the considerable number of Out Of Vocabulary (OOV) words offered by this language model over both DEV and TEST data sets must be taken into consideration along our experiments.

\subsection{Tool Setup}

As we have commented in Section \ref{asr}, the configuration of our recognition system is mainly based on the WSJ recipe. For the training phase, default parameters were kept. For decoding, we set a value of 13.0 to pruning beam and 6.0 to lattice beam. The language model covers scale factors between 1 and 20, while the speech model scale factor has a value of 0.08333. On the other hand, based on the BABEL recipe\footnote{\url{https://github.com/kaldi-asr/kaldi/blob/master/egs/babel/s5/local/score\_combine.sh}}, word insertion penalty values between -5.0 and 5.0 were studied. All these decoding parameters were evaluated in each experimental trial but only the lowest word error rate was considered.

\subsection{Evaluation}

All the results presented along our experiments are evaluated by the well-known Word Error Rate (WER) with 95\% confidence intervals obtained by the bootstrap method as described in \cite{bisani2004bootstrap}.

\section{LIP-RTVE Baseline Performance} \label{baseline}

A context-dependent GMM-HMM system, whose details are described in Section \ref{asr}, was employed in our baseline experiments. Different modalities, as Tables \ref{tab:audio-only} and \ref{tab:video-only} reflect, were studied over the data sets defined in Section \ref{datasets}. Concretely, we report the recognition performance obtained over the DEV and TEST sets along the training phases for both a SD and SI scenario.

Regarding our audio-only experiments, henceforth considered as the lower error bound for our task, we employed the 39-dimensional MFCCs (described in Section \ref{mfccs}) and the standard three-state HMM's topology. From the results reported in Table \ref{tab:audio-only}, the first aspect we must highlight is how, in both scenarios, the system recognition performance improves as we progress through the training stages; especially, since we build the first context-dependent system (DELTAS). Another aspect we must consider is that, as it might be expected, results for the SD scenario provide better recognition rates than those for the SI scenario.

\begin{table}[H]
\centering
\begin{adjustbox}{width=\columnwidth}
\begin{tabular}{ccccccc}
\toprule
\multicolumn{2}{c}{\multirow{2}{*}{\textbf{Dataset}}} & \multicolumn{4}{c}{\textbf{Training phases}} \\\cmidrule{3-6}
& & \textbf{MONO} & \textbf{DELTAS} & \textbf{LDA+MLLT} & \textbf{SAT} \\\midrule
\multirow{2}{*}{\textbf{SI}} & \textbf{DEV} & 40.7$\pm$1.1 & 20.9$\pm$0.9 & 18.8$\pm$0.9 & 16.9$\pm$0.8 \\
 & \textbf{TEST} & 40.4$\pm$1.2 & 20.0$\pm$0.9 & 16.7$\pm$0.9 & 15.3$\pm$0.8 \\\midrule
\multirow{2}{*}{\textbf{SD}} & \textbf{DEV} & 38.9$\pm$1.0 & 14.2$\pm$0.7 & 11.5$\pm$0.6 & 9.5$\pm$0.6 \\
 & \textbf{TEST} & 37.5$\pm$1.1 & 12.2$\pm$0.6 & 10.1$\pm$0.5 & 8.0$\pm$0.5 \\
\bottomrule
\end{tabular}
\end{adjustbox}
\caption{Audio-only baseline results (WER) for each training phase in both a speaker-independent (SI) and speaker-dependent (SD) scenario.}
\label{tab:audio-only}
\end{table}

\begin{figure}[!htbp]
  \centering
  \includegraphics[scale=0.17]{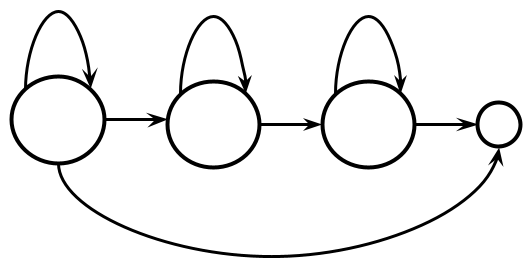}
  \caption{The HMM's topology employed in video-only experiments.}
  \label{topos}
\end{figure}

With respect to video-only experiments, the eigenlips described in Section \ref{visualfeats} were employed. Nevertheless, we first must consider that visual data presents a lower sample rate than audio data. Therefore, our first experiments were focused on defining the optimal HMM's topology, either adding transitions and/or reducing the number of states. Thus, the topology depicted in Figure \ref{topos} was found as the better approach to fit the temporary nature of our visual data.

When dealing with the VSR task, as it was mentioned in Section \ref{challenges}, their inherent challenges must be taken into consideration. Thus, as we can see in Table \ref{tab:video-only}, the SD scenario reflects, although the recognition rates are not comparable, an evolution similar to that observed in the audio-only experiments, where as we progress through the training phases, the error rate decreases. However, SI experiments did not reach acceptable results. This behaviour is in accordance with the work carried out by \newcite{cox2008challenge}, where the authors studied the different challenges posed by the SI setting. In any case, our results demonstrate that further research is necessary in the VSR task, either improving the quality of the extracted visual features or employing more powerful automatic systems.

\begin{table}[H]
\centering
\begin{adjustbox}{width=\columnwidth}
\begin{tabular}{ccccccc}
\toprule
\multicolumn{2}{c}{\multirow{2}{*}{\textbf{Dataset}}} & \multicolumn{4}{c}{\textbf{Training phases}} \\\cmidrule{3-6}
& & \textbf{MONO} & \textbf{DELTAS} & \textbf{LDA+MLLT} & \textbf{SAT} \\\midrule
\multirow{2}{*}{\textbf{SI}} & \textbf{DEV} & 96.5$\pm$0.3 & 95.9$\pm$0.2 & 96.0$\pm$0.3 & 95.9$\pm$0.3 \\
 & \textbf{TEST} & 96.5$\pm$0.4 & 96.2$\pm$0.2 & 96.3$\pm$0.3 & 95.9$\pm$0.2 \\\midrule
\multirow{2}{*}{\textbf{SD}} & \textbf{DEV} & 96.0$\pm$0.3 & 90.4$\pm$0.7 & 88.0$\pm$0.8 & 82.9$\pm$1.1 \\
 & \textbf{TEST} & 95.6$\pm$0.2 & 90.1$\pm$0.7 & 87.5$\pm$0.8 & 81.4$\pm$1.2 \\
\bottomrule
\end{tabular}
\end{adjustbox}
\caption{Video-only baseline results (WER) for each training phase in both a speaker-independent (SI) and speaker-dependent (SD) scenario.}
\label{tab:video-only}
\end{table}

Finally, regarding the audiovisual approach, different feature fusion methods \cite{potamianos2003recent} were explored. Nonetheless, these experiments did not improve the quality of the obtained audio-only results.


\section{Conclusions}

This paper has described a new audiovisual database to deal with unconstrained natural Spanish, following the so-called \textsl{in the wild} philosophy. More precisely, the compiled LIP-RTVE database offers around 13 hours of data semi-automatically collected from Spanish broadcast television. Thus, our contribution attempts to cover the relative lack of \textsl{in the wild} Spanish resources \cite{zadeh2020moseas} and the increased interest in VSR in recent decades \cite{fernandez2018survey}, since the LIP-RTVE database was primarily conceived to address this challenging task. On the other hand, in order to establish an experimental benchmark, both a SI and SD partition were defined. Baseline performances were obtained by employing the traditional GMM-HMM paradigm \cite{gales2008application}.

Regarding future work, we must first address the public distribution of our corpus, which will be released as soon as possible in a format that respects the NDA license of its source data. Once this issue is solved, we consider organizing competitions in order to encourage research on our database, especially in the VSR task which remains an open problem \cite{fernandez2018survey}. On the other hand, our future research is focused on exploring more robust and accurate systems where Deep Learning techniques would be incorporated. Concretely, in addition to study the combination of HMMs with Deep Neural Networks \cite{hinton2012deep}, we consider experimenting with end-to-end architectures \cite{chan2015listen,chung2017lip,afouras2018deep,ma2022visual}, whose possible benefits were indicated in Section \ref{rois}. Finally, we plan to increase the size of the LIP-RTVE database.

\section*{Acknowledgements}
This work was partially supported by Generalitat Valenciana under project DeepPattern (PROMETEO/2019/121) and by Ministerio de Ciencia under project MIRANDA-DocTIUM (RTI2018-095645-B-C22).

\section*{Bibliographical References}\label{reference}

\bibliographystyle{lrec2022-bib}
\bibliography{lrec2022-example}

\section*{Language Resource References}
\label{lr:ref}
\bibliographystylelanguageresource{lrec2022-bib}
\bibliographylanguageresource{languageresource}

\end{document}